\documentclass[10pt,twocolumn,letterpaper]{article}

\usepackage{cvpr}      % To produce the REVIEW version
\usepackage[ruled,linesnumbered]{algorithm2e}
\usepackage{multicol}
\usepackage{multirow}
\usepackage{bm}
\usepackage{amsmath} 
\usepackage{bbding}
\usepackage{pifont}
\usepackage{colortbl}

\definecolor{cvprblue}{rgb}{0.21,0.49,0.74}
\usepackage[pagebackref,breaklinks,colorlinks,allcolors=cvprblue]{hyperref}

\def\paperID{2970} % *** Enter the Paper ID here
\def\confName{CVPR}
\def\confYear{2025}

\title{AdvAnchor: Enhancing Diffusion Model Unlearning with Adversarial Anchors}

\author{Mengnan Zhao, Lihe Zhang, Xingyi Yang, Tianhang Zheng, Baocai Yin}

\begin{document}
\maketitle
\begin{abstract}
Security concerns surrounding text-to-image diffusion models have driven researchers to unlearn inappropriate concepts through fine-tuning. 
Recent fine-tuning methods typically align the prediction distributions of unsafe prompts with those of predefined text anchors.
However, these techniques exhibit a considerable performance trade-off between eliminating undesirable concepts and preserving other concepts.
In this paper, we systematically analyze the impact of diverse text anchors on unlearning performance. 
Guided by this analysis, we propose AdvAnchor, a novel approach that generates adversarial anchors to alleviate the trade-off issue.
These adversarial anchors are crafted to closely resemble the embeddings of undesirable concepts to maintain overall model performance, while selectively excluding defining attributes of these concepts for effective erasure. Extensive experiments demonstrate that AdvAnchor outperforms state-of-the-art methods. Our code is publicly available at \textcolor{black}{\url{https://anonymous.4open.science/r/AdvAnchor}}.
\end{abstract}

\section{Introduction}
\label{sec:intro}

Text-guided diffusion models (DMs) have garnered significant interest in the research community for their ability to generate high-fidelity images \cite{ song2020denoising, ho2022classifier} and their widespread applications like medical image reconstruction \cite{yang2023diffusion,kazerouni2022diffusion}, creative arts \cite{li2023loftq}, and material generation \cite{akrout2023diffusion,kingma2024understanding,yu2023diffusion}. 
However, these models also encounter critical security concerns, such as harmful content generation \cite{qu2023unsafe} and potential copyright infringement \cite{samuelson2023generative}. 
Furthermore, retraining safe DMs is challenging due to the extensive data cleaning, the high training resource consumption, and the unpredictable effects of training data on model predictions.

To remove unsafe behaviors from DMs without retraining, researchers are increasingly exploring Machine Unlearning (MU) techniques. A prominent MU approach directly fine-tunes the weights of pre-trained DMs \cite{gandikota2023erasing}, often by minimizing the prediction differences between text anchors (pre-defined target prompts) and prompts containing unsafe concepts \cite{zhang2023forget,gandikota2024unified}. 
For instance, to erase the `\textit{Van Gogh}' style, Abconcept \cite{kumari2023ablating} adjusts the cross-attention module weights in DMs to enhance prediction consistency between predefined prompt pairs, such as ``A picture of a painting" and ``A picture of a \textit{Van Gogh's} painting".
%\yxyc{Anchor is a key concept in the paper, but has never been properly defined in this context. Suggest to have a clear definition what a anchor mean before using it. For example, anchor is pre-defined target prompt. } 

% such as constructing evaluation datasets \cite{zhang2024unlearncanvas}, as well as white-box \cite{zhang2023generate,chin2023prompting4debugging} and black-box \cite{tsai2023ring} prompt debugging. Additionally, the erasing transferability is also explored \cite{lyu2023one}.

%approaches investigating erasing transferability are being explored \cite{lyu2023one}.
%  and preserving model generation capability on regular concepts during concept unlearning \cite{huang2023receler,hong2023all}. 
%  and the 

\begin{figure}[t]
    \begin{center}
        %\fbox{\rule{0pt}{2in} \rule{0.9\linewidth}{0pt}  }
        \includegraphics[width=1\linewidth]{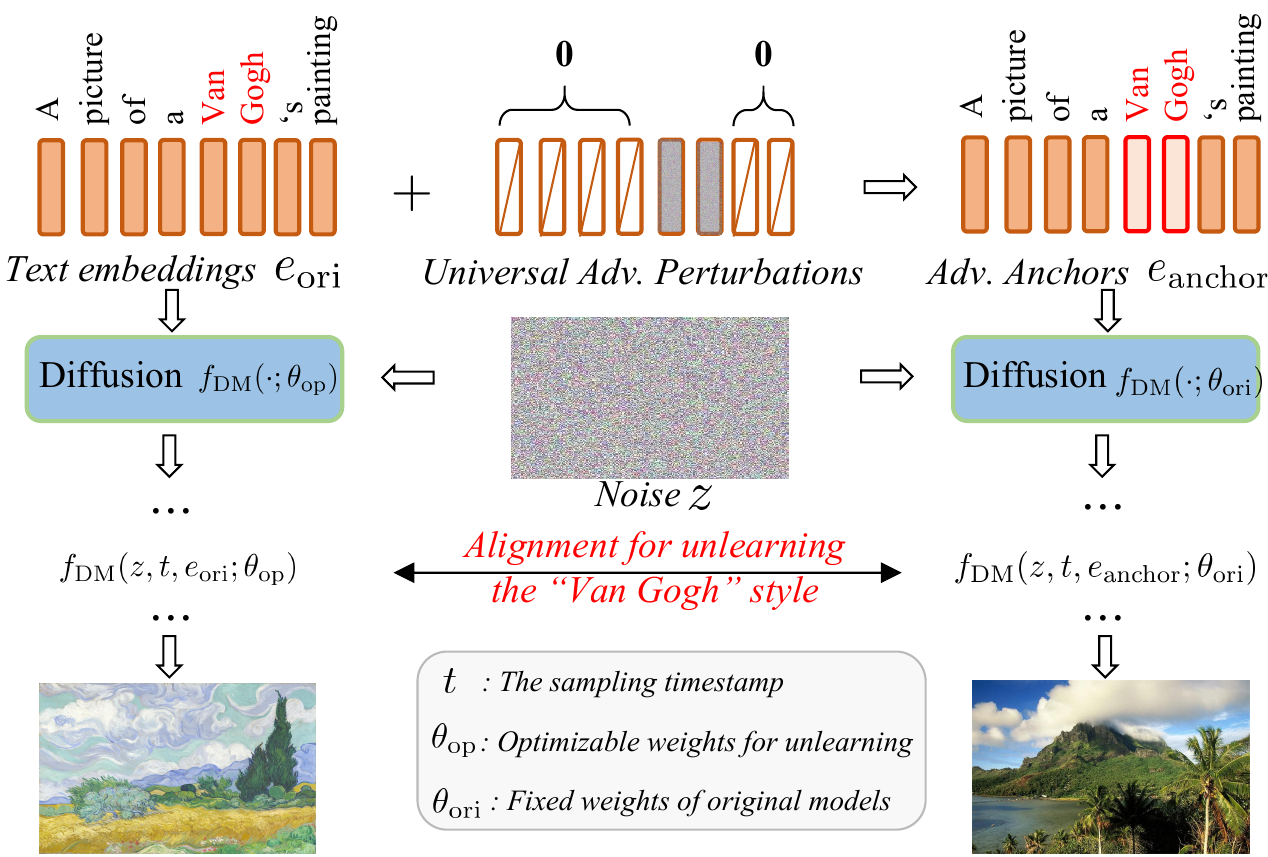}
    \end{center}
    \vspace{-2mm}
    \caption{    
        Overview of the proposed AdvAnchor. To construct adversarial anchors, tiny adversarial perturbations that greatly affect the generation performance of DMs on `\textit{Van Gogh}' are added to the embeddings of `\textit{Van Gogh}'.
\(\bm{\theta}_\text{op}\) is fine-tuned by aligning the prediction distributions of \(\bm{e}_\text{ori}\) with those of \(\bm{e}_\text{anchor}\).
    }
    % \vspace{-2mm}
    \label{fig1}
\end{figure}

While recent MU methods effectively erase undesirable concepts from pre-trained DMs, they often significantly compromise the overall generation quality of models. 
For instance, after erasing the `\textit{Cezanne}' style with Abconcept, the model struggles to yield images in the styles of `\textit{Van Gogh}' or `\textit{Picasso}'. 
Inspired by prior works \cite{gandikota2023erasing,kim2023towards} that utilize different anchors to fine-tune model weights, we naturally raise a key question: 
\textit{How does anchor selection influence DM unlearning,  and could performance be enhanced through improved anchor crafting?}

To answer this question, we conduct a systematic experimental analysis on the impact of anchors in DM unlearning. 
Our findings have twofold: 
1) Anchors that closely resemble the undesirable concept are more effective in preserving overall model performance;
2) Effective erasure occurs when anchors omit the defining attributes of the undesirable concept. Namely, for effective erasure with minimal impact on model performance, ideal anchors should approximate the undesirable concept yet exclude its defining attributes. 

Based on our analyses and prior studies (tiny adversarial perturbations can substantially affect model predictions \cite{goodfellow2014explaining,zhao2024adversarial}), 
we introduce AdvAnchor, a simple yet effective adversarial anchoring method for DMs.
AdvAnchor creates adversarial anchors by adding crafted universal perturbations to the embeddings of undesirable concepts. 
These perturbations, guided by the proposed similarity losses and optimization strategies, aim to degrade the generation quality of undesirable concepts across any visual input, thereby removing defining attributes of these concepts.
Notably, the adversarial anchors are versatile and can be integrated into various MU techniques, such as the alignment mechanism in Abconcept. The AdvAnchor pipeline is shown in Fig. \ref{fig1}.

% This paper primarily contains three contributions:
Our contributions are three-fold:
{\bf (1)} We conduct a systematic analysis of anchor impact on DM unlearning, revealing that for effective concept erasure with minimal impact on model performance, ideal anchors should maintain semantic similarity to the undesirable concept while excluding its defining attributes.
{\bf (2)} Based on this insight, we propose AdvAnchor, a simple yet effective method that generates adversarial anchors using designed loss constraints and optimization strategies.
{\bf (3)} %Extensive experiments demonstrate that AdvAnchor significantly mitigates the trade-off between the efficacy of concept erasure and the preservation of model performance.
Extensive experiments demonstrate that AdvAnchor significantly enhances both erasure and preservation performance in DM unlearning.

\section{Related work}
\label{sec:formatting}
\begin{figure}[t]
    \begin{center}
        %\fbox{\rule{0pt}{2in} \rule{0.9\linewidth}{0pt}  }
        \includegraphics[width=1\linewidth]{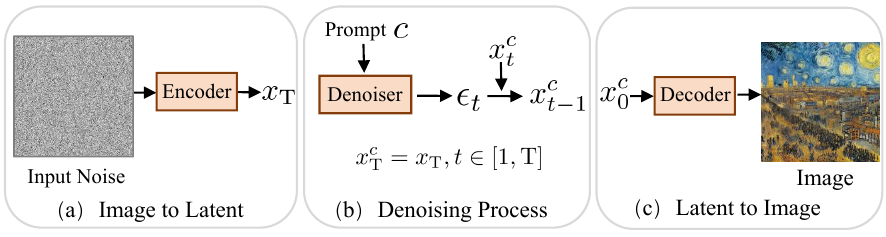}
    \end{center}
    \vspace{-2mm}
    \caption{    
        The denoising process of text-guided DMs.
        a) the encoder converts the input noise into latent representations \(\bm{x}_\text{T}\); b) the denoiser iteratively removes the predicted noise \(\bm{\epsilon}_{t\in[1,\text{N}]}\) from latent representations \(\bm{x}_t^c\); c) the decoder reconstructs the image from the denoised representations \(\bm{x}_0^c\).
    }
    \label{fig2}
    % \vspace{-2mm}
\end{figure}

% which involve a two-step process: first introducing noise to convert the %data distribution into a Gaussian distribution, followed by denoising to restore the original distribution.   
% These encompass the risk of copyright infringement , exacerbation of social biases \cite{liu2023geom}, and the potential for generating inappropriate content \cite{yang2023mma}. 
% and comply with regulatory requirements such as the ``Right to Be Forgotten" \cite{rosen2011right}, 
% MU techniques are applied into DMs to eliminate specific concepts from pre-trained models through fine-tuning.
Recent advancements in text-to-image generation \cite{ramesh2021zero,zhou2022towards}, especially with the emergence of DMs \cite{ramesh2022hierarchical,rombach2022high}, have drastically improved the quality of  high-resolution image generation \cite{shivashankar2023semantic,kawar2023imagic}.
However, these developments also raise security issues
\cite{somepalli2023diffusion}, such as the dissemination of NSFW materials \cite{tsai2023ring,liu2023geom} and copyright infringement \cite{shan2023glaze}.
To mitigate these risks,
the MU fine-tuning technique \cite{bourtoule2021machine, gupta2021adaptive,chourasia2023forget} has been proposed to eliminate specific concepts from pre-trained models.
In this work, we focus on the DM unlearning task.
Figure~\ref{fig2} illustrates the denoising process in text-guided DMs, which consists of two-steps: introducing noise to convert the data distribution into a Gaussian distribution, followed by denoising to restore the original distribution.

Several advanced MU techniques have been developed for DMs.
For instance, Forget-Me-Not \cite{zhang2023forget} eliminates undesirable concepts by decreasing the attention map values associated with these concepts. 
GEOM \cite{liu2023geom} introduces implicit concept erasure, focusing on removing concepts that cannot be controlled through pre-defined text prompts.
P4D \cite{chin2023prompting4debugging} and UnlearnDiff \cite{zhang2023generate} evaluate the robustness of unlearning approaches by generating adversarial text prompts \cite{zhang2024unlearncanvas}.
Receler \cite{huang2023receler} improves the erasing robustness by introducing an adversarial learning strategy, \textit{i.e.}, expanding the training data with adversarial soft prompts.

Particularly, depending on whether the network structure is modified, DM unlearning techniques can be categorized into adaptor-based and adaptor-free approaches. The former modifies the DM structure by introducing adaptor layers, learning only their weights.
For instance, SPM \cite{lyu2023one} constructs lightweight one-dimensional adapters \cite{chin2023prompting4debugging,mehrabi2023flirt} and MACE \cite{lu2024mace} integrates multiple non-interfering LoRA modules \cite{hu2021lora}. 
In contrast, the latter directly updates the module weights in original DMs.
For example, ESD \cite{gandikota2023erasing} and Abconcept \cite{kumari2023ablating} fine-tune cross-attention modules, while SepME \cite{zhao2024separable} restricts weight modifications to image-independent layers.
Our approach follows the adaptor-free paradigm, \textit{i.e.}, fine-tuning the cross-attention modules of DMs.
% In contrast, the latter directly modifies the weights of original models by XXX. %zlh  请补充
% zmn by xxx这一部分是在下一段讲的，这里其实解释的是为什么不用adapter-based方法作为baseline，不涉及adapter-free的具体实现。
%Particularly, according to the location of learnable weights, the fine-tuning techniques can be categorized into two manners: original model fine-tuning and plug-in based fine-tuning.
%The former directly modifies the original model weights. 
%In contrast, the plugin-based approach modifies the model structure by introducing additional components and only learns the parameters of these introduced components.
% In contrast,  the plug-in based approaches disentangle concepts from pre-trained models.
%For instance, SPM \cite{lyu2023one} constructs lightweight one-dimensional adapters \cite{chin2023prompting4debugging,mehrabi2023flirt} and MACE \cite{lu2024mace} integrates multiple non-interfering LoRA modules \cite{hu2021lora}. 
%

In this paradigm, researchers generally align the prediction distributions for undesirable concepts with those for predefined anchors. 
For instance, SDD \cite{kim2023towards} employs an empty prompt as the anchor for all undesirable concepts. Abconcept utilizes broader concepts compared to undesirable concepts as anchors.
ESD uses both the undesirable concept and the empty prompt to create anchors.
UCE \cite{gandikota2024unified} generates anchors based on a weighted combination of various text prompts, aiming to address generation bias.
All-but-one \cite{hong2023all} presents a Prompt-to-Prompt erasing technique that employs the highest and lowest embedding values of target prompts \cite{brack2024sega} to construct anchors.
% All-but-one \cite{hong2023all} employs a Prompt-to-Prompt erasing technique that leverages the highest and lowest embeddings of target prompts \cite{brack2024sega} to construct anchors.
% most representative features predicted from the specified text anchor. These features are primarily found in  pixel values .
%
% Unlike these techniques that construct anchors based on text prompts, this paper aims to comprehensively understand the impact of diverse anchors on the unlearning performance and generate adversarial anchors through optimization.
Unlike these methods that combine predefined text prompts to build anchors, this paper systematically explores the anchor impact on unlearning performance and introduces adversarial anchors.

\section{Proposed method}
In this section, we first explore the impact of anchors on DM unlearning. Then, we introduce the proposed AdvAnchor.
% the experimental analysis is provided.
% Subsequently, we analyze the performance disparity introduced by diverse anchors to gain a deeper understanding of DM unlearning. 
% Finally, we illustrate the proposed adversarial anchoring method.

\begin{figure*}[t]
    \begin{center}
        %\fbox{\rule{0pt}{2in} \rule{0.9\linewidth}{0pt}  }
        \includegraphics[width=0.9\linewidth]{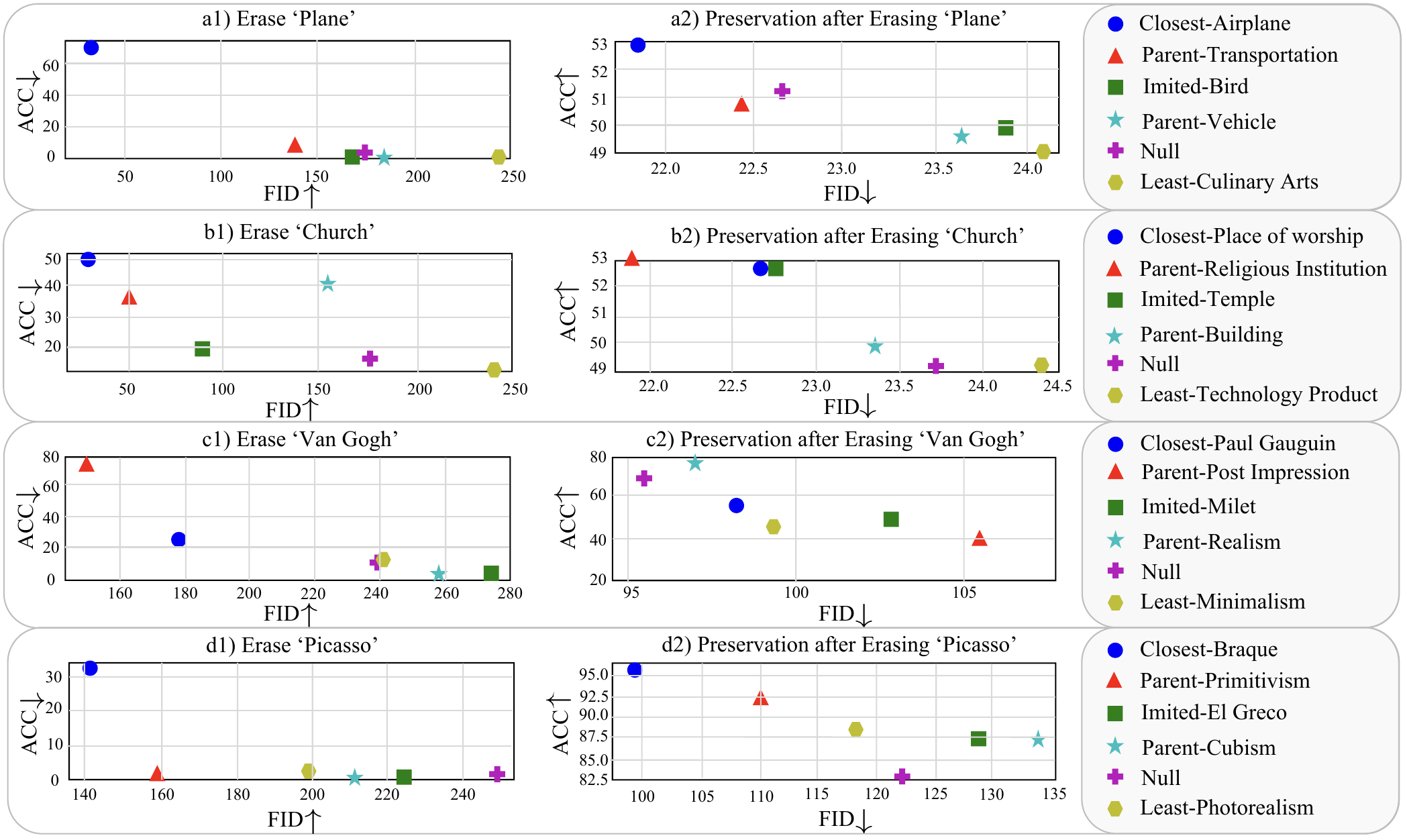}
    \end{center}
    \vspace{-4mm}
    \caption{    
 Impact of using various types of words as anchors (\(p_\text{anchor}^\text{word}\)) on DM unlearning.
    }
    \label{fig3}
    % \vspace{-2mm}
\end{figure*}

\subsection{Impact of various anchors on DM unlearning}\label{sec3.1}
% To investigate the impact of anchors on DM unlearning, we adopt the popular MU paradigm , 
We follow previous unlearning approaches  \cite{kim2023towards,gandikota2023erasing} to conduct this research, \textit{i.e.}, the undesirable concept \(c_\text{u}\) is erased from DMs by minimizing the prediction difference between \(p_\text{u}\) and \(p_\text{anchor}\). Here, \(p_\text{u}\) is a prompt containing \(c_\text{u}\) and \(p_\text{anchor}\) denotes a target prompt. 
\begin{equation}\label{eq1}
% \small
\min_{\bm{\theta}_\text{op}} \mathcal{L}_\text{op} = 
        \|f_\text{de}(\bm{x}_t, \bm{e}_\text{pu}; \bm{\theta}_\text{op}) - f_\text{de}(\bm{x}_t,\bm{e}_\text{anchor};\bm{\theta}_\text{ori})\|_2,
\end{equation}
where \(\bm{\theta}_\text{op}\) denotes the optimizable model weights for unlearning. 
\(f_\text{de}(\cdot)\) means the denoiser in Fig.~ \ref{fig2}.
\(\bm{x}_t\) represents the latent representations of inputs at the timestamp \(t\), which can be obtained through either the diffusion process \cite{gandikota2023erasing} or the sampling process \cite{kumari2023ablating}. 
\(\bm{e}_\text{pu}\) and \(\bm{e}_\text{anchor}\) are text embeddings of prompts \(p_\text{u}\) and \(p_\text{anchor}\), respectively.
% \(c_{gt}\) is the ground-truth concept of \(\bm{x}_t\).  
\(\bm{\theta}_\text{ori}\) refers to the fixed weights of original models.
\(\|\cdot\|_2\) is the \(\ell_2\) norm.
% Each data within the dataset \(\mathcal{D}_\text{u}\) contains the concept \(c_f\).
% For time efficiency, we yield \(\bm{x}_t\) within the diffusion process. 
\begin{figure*}[t]
    \begin{center}
        %\fbox{\rule{0pt}{2in} \rule{0.9\linewidth}{0pt}  }
        \includegraphics[width=0.9\linewidth]{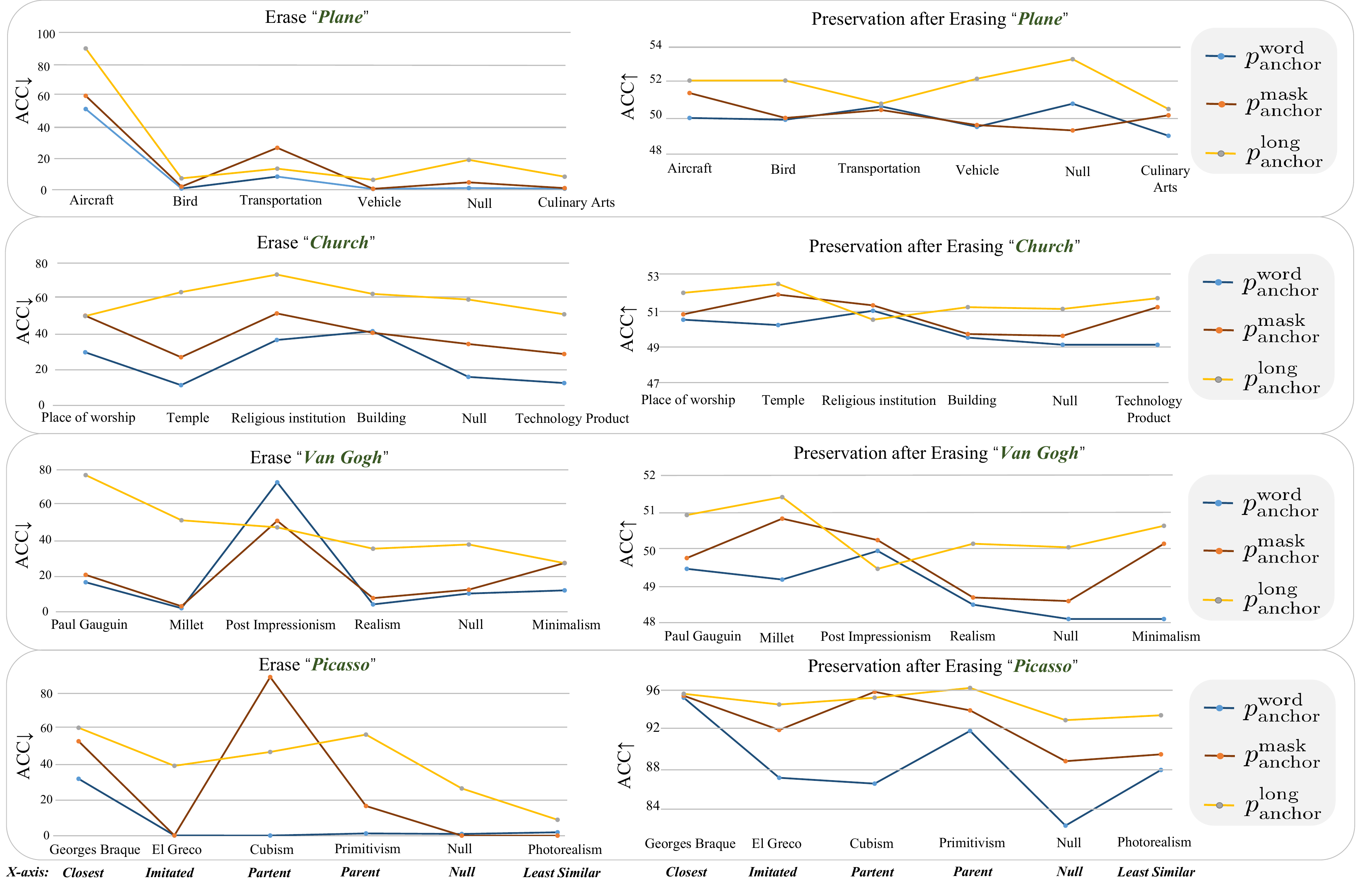}
    \end{center}
    \vspace{-4mm}
    \caption{    
        Ablation studies on the length of the shared sentence between \(p_\text{anchor}\) and \(p_\text{u}\) in DM unlearning.
    }
    \label{fig4}
\end{figure*}

\subsubsection{Anchors} 
% The following describes the types of \( p_{anchor}\).
% The types of \(p_\text{anchor}\) and their corresponding \(p_\text{u}\) are described below.
% We establish several fixed criteria to generate anchors \( p_{anchor} \) for each undesirable concept \(c_\text{u}\):
We use various types of \(p_\text{u}\) and \(p_\text{anchor}\), detailed as follows:

% {1) Each anchor is designated as a word, termed as \(p_\text{anchor}^\text{word} \), and \(p_\text{u} = c_\text{u}\). We employ ChatGPT to determine the anchor of \( c_\text{u} \), and the input prompts may be as follows:
{1) Each anchor is designated as a word, denoted \( p_\text{anchor}^\text{word} \), with \(p_\text{u} = c_\text{u}\). ChatGPT \cite{achiam2023gpt} is used to determine the anchor for \( c_\text{u} \), with potential input prompts structured as follows:
\begin{itemize}[leftmargin=2em]
\setlength{\parskip}{0pt}
    \item[-] \textit{The closest type to \( c_\text{u} \)};
    \item[-] \textit{The type imitated by \( c_\text{u} \)};
    \item[-] \textit{The parent class of \( c_\text{u} \)};
    \item[-] \textit{The type least similar to \( c_\text{u} \)}.
\end{itemize}

2) We embed \(p_\text{anchor}^\text{word} \) and \(c_\text{u}\) into a masked sentence to construct \(p_\text{anchor}^\text{mask} \) and \(p_\text{u}\), respectively.
For instance, when removing an artist style from DMs,  \(p_\text{anchor}^\text{mask} \) and \(p_\text{u}\) can be denoted as ``\textit{A picture of a \(\{p_\text{anchor}^\text{word}\} \)'s painting}" and ``\textit{A picture of a \(\{c_\text{u}\}\)'s painting}", respectively.

3) We prepend a same long sentence to \(p_\text{anchor}^\text{word}\) and \(c_\text{u}\), to construct \(p_\text{anchor}^\text{long}\) and \(p_\text{u}\), respectively. For example, when erasing the `\textit{Van Gogh}' style, the prompt for ChatGPT to generate this long sentence might be:
\begin{itemize}[leftmargin=2em]
\setlength{\parskip}{0pt}
\item[-] 
\textit{Provide a long sentence that describes an artistic style except for `Van Gogh'}.
\end{itemize}

4) We utilize a descriptive sentence as an anchor, called as \(p_\text{anchor}^\text{desc}\), and \(p_\text{u} = c_\text{u}\).
For instance, when erasing an object category, the prompt for ChatGPT may
 be 
% \begin{itemize}
% [leftmargin=2em]
% \setlength{\parskip}{0pt}
% \item[-] 

\(p_\text{anchor}^\text{desc\_1}\): \textit{Provide a sentence that describes the morphological features of \(p_\text{anchor}^\text{word} \). This sentence should include the common attributes between \(p_\text{anchor}^\text{word} \) and \(c_\text{u}\)}.
% \item[-] 

\(p_\text{anchor}^\text{desc\_2}\):
\{\(p_\text{anchor}^\text{desc\_1}\)\}, 
\textit{
% Provide a sentence that describes the morphological features of \(p_\text{anchor}^\text{word} \). This sentence should include the common attributes between \(p_\text{anchor}^\text{word} \) and \(c_\text{u}\), 
% and the unique key morphological characteristics of \(p_\text{anchor}^\text{word} \), 
while excluding defining features of \(c_\text{u}\)}.
% and the unique key morphological characteristics of \(p_\text{anchor}^\text{word} \), 
% while   features specific to \(c_\text{u}\)}.
% \end{itemize}

% \subsubsection{Empirical Analysis.} 
% 这一部分主要包括评估指标，评估数据，以及实验的细节部分，我不太清楚能不能使用Empirical Analysis
% Empirical Analysis这个究竟应该指代哪一部分？
\subsubsection{Settings}\label{settings}

\textit{Evaluation metrics.} 
1) We use Fréchet Inception Distance (FID) \cite{heusel2017gans} to measure the distance between images generated with the undesirable concept and the corresponding anchor;
2) Accuracy (ACC): For object classification, we use a pre-trained ResNet50 \cite{he2016deep}. For style classification, we fine-tune the fully connected layer of a pre-trained ResNet18 \cite{he2016deep} on a dataset generated by original DMs. This dataset includes a blank prompt and nine artist styles: \textit{Cezanne, Van Gogh, Picasso, Jackson Pollock, Caravaggio, Keith Haring, Kelly McKernan, Tyler Edlin}, and \textit{Kilian Eng}.
% \end{itemize}

\textit{Evaluation data.} 
% We evaluate the erasure and preservation performance of each unlearned model.
% \begin{itemize}
% [leftmargin=2em]
% \setlength{\parskip}{0pt}
% \item[-] 
1) Erasure: Generate 1,000 images per undesirable concept, with 200 seeds per concept and 5 images per seed.
% \item[-] 
2) Object preservation: Generate 1,859 images with prompts from 1,000 categories in the ImageNet dataset\footnote{\url{https://github.com/rohitgandikota/erasing}}.
% \item[-] 
3) Style preservation: Generate 2,000 images with the retained styles as prompts (excluding the erased one), with 50 seeds per style and 5 images per seed.
% \end{itemize}

\textit{Others.} 
Under identical settings, models from multiple fine-tuning processes exhibit notable performance variation. 
Hence, we report the average evaluation results over three independent fine-tuning runs for each undesirable concept.
% , with all comparative experiments conducted with the same number of fine-tuning steps.
% Besides, all experiments are conducted with the same number of fine-tuning steps.

\subsubsection{Observations}
The observations are as follows:
\begin{itemize}[leftmargin=2em]
\setlength{\parskip}{0pt}
\item[-o1] A higher similarity between \(p_\text{anchor}^\text{word}\) and \(c_\text{u}\) often results in better preservation performance, as shown in Fig.~\ref{fig3}.
\item[-o2] A longer shared sentence between \(p_\text{anchor}\) and \(p_\text{u}\) generally leads to better preservation performance, as demonstrated in Fig.~\ref{fig4}, which compares the results for \(p_\text{anchor}\) set to \(p_\text{anchor}^\text{word}\), \(p_\text{anchor}^\text{mask}\), and \(p_\text{anchor}^\text{long}\).
\item[-o3] Unlearning with \(p_\text{anchor}^\text{desc\_2}\) usually achieves superior erasure and preservation performance than other variants, as illustrated in Fig.~ \ref{fig5}. 
\end{itemize}

\begin{figure*}[t]
    \begin{center}
        %\fbox{\rule{0pt}{2in} \rule{0.9\linewidth}{0pt}  }
        \includegraphics[width=0.9\linewidth]{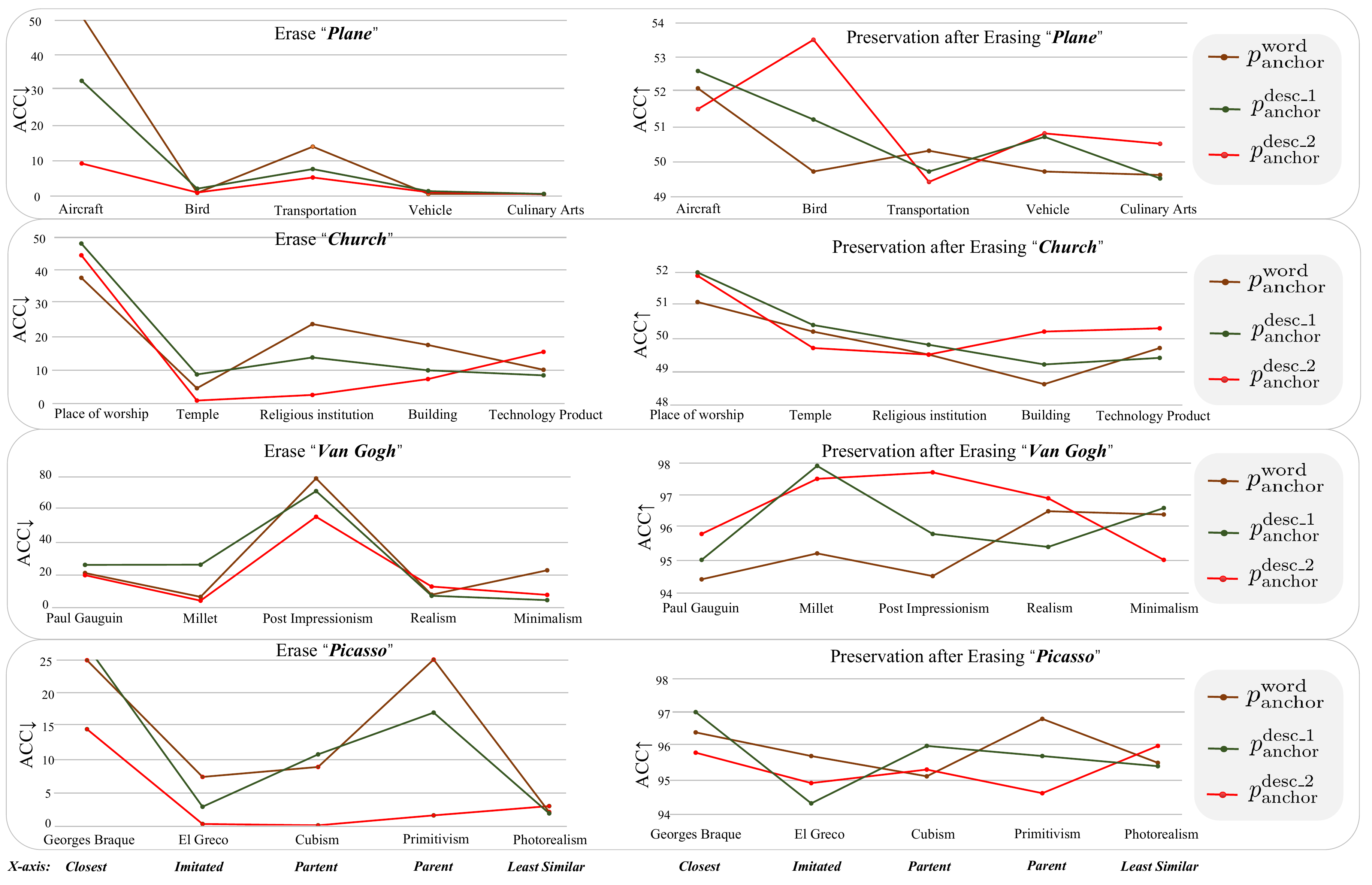}
    \end{center}
    \vspace{-4mm}
    \caption{    
         Comparative experiments using \(p_\text{anchor}^\text{word}\) and \(p_\text{anchor}^\text{desc}\).
    }
    \label{fig5}
    \vspace{-2mm}
\end{figure*}

% These findings indicate the substantial impact of anchors on DM unlearning:
Additional experiments supporting these observations are provided in the Appendix. We can conclude that:
\begin{itemize}
% [leftmargin=2em]
\setlength{\parskip}{0pt}
 \item[-] \textit{To preserve overall model performance, \(\bm{e}_\text{anchor}\) and \(\bm{e}_\text{pu}\) should retain high similarity (Observations {o1} and {o2})}.
\item[-] \textit{Undesirable concepts can be 
% effectively 
erased by excluding their defining attributes from anchors (Observation {o3}).}
\end{itemize}

% }

% Inspired by the strategy of identifying the nearest misclassification class in classification tasks, we introduce AdvAnchor in MU to generate adversarial anchors that primarily contain defining attributes.

% \textit{Untargeted attack.} 
% Given a classification model \(\mathcal{M}\) and the ground truth class \(y\) for \(\bm{x}_0\), an adversarial perturbation \(\bm{\delta}\) is introduced to \(\bm{x}_0\). This perturbation is iteratively adjusted to maximize the loss \(\mathcal{L}(\mathcal{M}(\bm{x}_0 + \bm{\delta}), y)\). The process continues until the model \(\mathcal{M}\) misclassifies the perturbed image, such that \(\arg \max \mathcal{M}(\bm{x}_0 + \bm{\delta}) \neq y\).

\subsection{Proposed AdvAnchor}
% We first illustrate the conventional untargeted attack \cite{madry2017towards}, and then describe the procedures for generating adversarial anchors.
Inspired by prior studies that tiny adversarial perturbations can significantly affect model predictions \cite{madry2017towards,xue2024diffusion}, and by our conclusions, we propose AdvAnchor to generate adversarial anchors for DM unlearning.

Specifically, as the text space is discrete, AdvAnchor fine-tunes the undesirable concept \(c_\text{u}\) in the embedding space to generate adversarial anchors \(\bm{e}_\text{anchor}^\text{adv}\).
The \(i\)-th feature element of \(\bm{e}_\text{anchor}^\text{adv}\)is expressed as
\[
\bm{e}_{\text{anchor},i}^\text{adv} = \left\{
\begin{array}{cc}
        \bm{e}_{\text{pu}, i} + \bm{e}_\text{adv}& \text{if } \bm{e}_{\text{pu}, i} = \bm{e}_{\text{u}} \\
        \bm{e}_{\text{pu}, i} & \text{otherwise},
\end{array}
\right.
\]
where \(\bm{e}_{\text{pu}}\) and \(\bm{e}_{\text{u}}\) denote the text embeddings of \(p_\text{u}\) and \(c_\text{u}\), respectively.
Since \(\bm{e}_{u}\) remains fixed during erasure, we designate \(\bm{e}_\text{adv}\) as a universal perturbation.

% aim to degrade the generation quality of undesirable concepts across any visual input, thereby 

To exclude defining attributes of undesirable concepts from \(\bm{e}_\text{anchor}^\text{adv}\), given the ground truth images \(\bm{x}_\text{gt}\) of undesirable concepts and the random noise \(\bm{z}\), we produce \(\bm{e}_\text{adv}\) by 
\begin{equation}\label{eq2sup}
    \max_{\bm{e}_\text{adv}} \|f_\text{DM}(\bm{z}, \bm{e}_\text{anchor}^\text{adv};\bm{\theta}_\text{ori}) - \bm{x}_\text{gt}\|_2,
\end{equation}
where \(f_\text{DM}\) denotes a pre-trained DM with fixed weights \(\bm{\theta}_\text{ori}\).
Eq. (\ref{eq2sup}) aims to degrade the generation quality of undesirable concepts across any visual input.
% ensures that the model output for \(\bm{e}_\text{anchor}^\text{adv}\) deviates from \(\bm{x}_\text{gt}\).
To influence the full denoising process with \(\bm{e}_\text{adv}\), we reformulate Eq. (\ref{eq2sup}) as:
\begin{equation}\label{eq2}
    \max_{\bm{e}_\text{adv}} [\mathcal{L}_\text{adv}(f_\text{de}(\bm{x}_t, \bm{e}_\text{anchor}^\text{adv};\bm{\theta}_\text{ori}), f_\text{de}(\bm{x}_t, \bm{e}_\text{pu};\bm{\theta}_\text{ori}))],
\end{equation}
% where \(\bm{x}_t\) denotes visual embeddings at a randomly selected timestamp \(t\), \(f_\text{de}(\cdot)\) means the denoiser in DMs, and 
where \(f_\text{de}(\bm{x}_t, \bm{e}_\text{pu};\bm{\theta}_\text{ori})\) acts as the pseudo target.
We design two functions for \(\mathcal{L}_\text{adv}\): \(\mathcal{L}_\text{adv1}\), which measures the cosine similarity between model predictions for \(\bm{e}_\text{anchor}^\text{adv}\) and \(\bm{e}_\text{pu}\),
\[
\max_{\bm{e}_\text{adv}} [- \cos(f_\text{de}(\bm{x}_t,\bm{e}_\text{anchor}^\text{adv}, \bm{\theta}_\text{ori}), f_\text{de}(\bm{x}_t,\bm{e}_\text{pu},\bm{\theta}_\text{ori}))],
\]
and \(\mathcal{L}_\text{adv2}\), which measures the similarity between generation changes \(g_c(\cdot)\) caused by \(\bm{e}_\text{anchor}^\text{adv}\) and \(\bm{e}_\text{pu}\):
\[
\max_{\bm{e}_\text{adv}} [- \cos(g_c(\bm{e}_\text{anchor}^\text{adv}), g_c(\bm{e}_\text{pu}))],\]
\[g_c(\bm{e}) = f_\text{de}(\bm{x}_t, \bm{e}, \bm{\theta}_\text{ori}) - f_\text{de}(\bm{x}_t, \bm{e}_\emptyset, \bm{\theta}_\text{ori}).
\]

Inspired by vector orthogonality, we set \(\mathcal{L}_{\text{adv}}(\cdot) = 0\) as the stopping criterion for optimizing \(\bm{e}_{\text{adv}}\).
The generated \(\bm{e}_\text{anchor}^\text{adv}\) can be easily integrated into various unlearning techniques by replacing predefined anchors, such as \(\bm{e}_\text{anchor}\) in Eq. (\ref{eq1}).
Furthermore, we design alternating, sequential, and cyclical optimization strategies to update \(\bm{\theta}_\text{op}\) and \(\bm{e}_{\text{adv}}\):

\begin{itemize}
\setlength{\parskip}{0pt}
    \item [-] \textit{Alternating optimization:} 
We \textit{alternately} update \(\bm{e}_{\text{adv}}\) and \(\bm{\theta}_{\text{op}}\) with various inputs.
For each input \(\bm{x}_t\), we stop optimizing \(\bm{e}_{\text{adv}}\) in Eq. (\ref{eq2}) when \(\mathcal{L}_{\text{adv}}(\cdot) = 0\), and then fine-tune \(\bm{\theta}_{\text{op}}\) using the current \(\bm{e}_{\text{adv}}\) and \(\bm{x}_t\).
    \item [-] \textit{Sequential optimization:} 
    We first construct the final \(\bm{e}_{\text{adv}}\) using various inputs, and then fine-tune \(\bm{\theta}_{\text{op}}\) with the final \(\bm{e}_{\text{adv}}\) across these inputs. Each input is \textit{repeatedly} used to update \(\bm{e}_{\text{adv}}\) until the stopping criterion is met.
    \item [-] \textit{Cyclical optimization:} 
    In contrast to sequential optimization, each input is revisited in \textit{cycles} to update \(\bm{e}_{\text{adv}}\) until the stopping criterion is met.
% We cyclically optimize \(\bm{e}_{\text{adv}}\) across different inputs \(\bm{x}_t\), followed by fine-tuning \(\bm{\theta}_{\text{op}}\) with the ultimate \(\bm{e}_{\text{adv}}\) and these inputs.
\end{itemize}

\begin{table*}[t]
    \tabcolsep = 0.04cm
    \begin{center}
    \caption{Comparative results on style unlearning. 
    \(c_{0\sim 8}\) represent nine artist styles in Section \ref{settings}.
    % `Cezanne', `VanGogh', `Picasso', `Jackson Pollock', `Caravaggio', `KeithHaring', `Kelly McKernan', `Tyler Edlin', `Kilian Eng', respectively.
    We evaluate both erasure (FID$\uparrow$/ACC$\downarrow$/LPIPS$\uparrow$) and preservation (FID$\downarrow$/ACC$\uparrow$/LPIPS$\downarrow$) performance.
    The red digits — \textbf{\textcolor{red!100}{digit$_1$}}, \textcolor{red!100}{\underline{digit$_2$}}, and \textcolor{red!50}{digit$_3$} — indicate the best, second-best, and third-best erasure results, respectively.
    Similarly, the \textcolor{blue!100}{blue digits} highlight best results for preservation performance.
    %while the blue digits — {digit$_1$}}, {digit$_2$}, and {digit$_3$} —  denote the top three preservation results in decreasing order.
    }
    % Digits in varying shades — \textcolor{black!100}{digit$_1$}}, \textcolor{black!100}{digit$_2$}, and \textcolor{black!30}{digit$_3$} — indicate the best, second-best, and third-best results, respectively.}
    \label{style}
    % \vspace{-1mm}
    \tiny
    \resizebox{1\textwidth}{!}{
    \begin{tabular}{ccc|cc ccccc ccccc}
    \toprule[1pt]
&\multicolumn{2}{c|}{ORI}&\multicolumn{2}{c}{SDD \cite{kim2023towards}}&\multicolumn{2}{c}{All-but-one \cite{hong2023all}}&\multicolumn{2}{c}{ESD \cite{gandikota2023erasing}}&\multicolumn{2}{c}{AbConcept \cite{kumari2023ablating}}&\multicolumn{2}{c}{\textbf{Ours}$(\mathcal{L}_\text{adv1})$}&\multicolumn{2}{c}{\textbf{Ours}$(\mathcal{L}_\text{adv2})$}\\
&Erase&Preserve&Erase&Preserve&Erase&Preserve&Erase&Preserve&Erase&Preserve&Erase&Preserve&Erase&Preserve\\
\midrule[0.5pt]

{\bf $c_0$}&0/98.0/0&0/99.4/0&282.1/1.20/.402&
108.5/87.8/.214&251.2/9.60/.371&103.6/91.2/.209&
320.3/0/.494&140.6/73.7/.278&213.4/18.8/.339&
115.0/85.1/.207&404.8/0/.385&96.4/94.3/.158&
261.6/22.0/.320&84.9/96.2/.141\\

{\bf  $c_1$}&0/97.6/0&0/99.4/0&255.3/12.4/.427&
95.2/95.8/.178&234.2/12.8/.445&97.9/95.6/.183&
278.4/4.80/.518&117.1/84.8/.220&246.7/34.8/.410&
75.7/96.9/.109&404.5/34.0/.522&113.0/96.0/.165&
256.0/31.2/.411&70.6/97.5/.090\\

{\bf $c_2$}&0/99.6/0&0/99.2/0&240.1/1.20/.422&
105.7/90.4/.189&261.0/0/.385&109.2/92.4/.189&
291.9/0/.494&149.2/71.4/.249&231.7/0.40/.356&
98.7/89.7/.157&452.0/0/.519&145.0/92.6/.181&
253.7/0/.380&96.6/91.2/.174\\

{\bf $c_3$}&0/99.6/0&0/99.2/0&424.5/4.40/.580&
91.6/93.7/.192&345.9/21.2/.414&88.1/95.2/.188&
469.9/0/.731&103.2/86.6/.229&317.0/5.60/.447&
69.4/97.0/.118&585.3/0/.386&108.8/94.8/.210&
338.4/5.20/.436&70.6/95.8/.128\\

{\bf  $c_4$}&0/99.6/0&0/99.2/0&302.9/0.80/.358&
98.5/94.0/.171&290.8/19.2/.298&100.5/95.2/.182&
321.3/1.60/.366&110.6/92.9/.186&243.9/17.2/.306&
80.6/95.5/.122&509.4/0/.600&123.3/90.9/.188&
283.1/11.6/.297&74.6/95.9/.117\\

{\bf $c_5$}&0/98.8/0&0/99.3/0&258.8/6.40/.605&
96.5/94.3/.164&238.5/9.20/.566&94.8/96.6/.151&
295.6/0.40/.635&102.4/91.6/.169&271.2/2.80/.621&
79.0/95.4/.105&309.0/0/.598&97.2/94.0/.160&
276.4/0/.648&74.4/94.9/.097\\

{\bf $c_6$}&0/100.0/0&0/99.0/0&242.2/1.60/.367&
99.8/93.2/.187&254.7/2.80/.433&99.7/95.3/.193&
286.6/0/.482&107.6/93.9/.189&174.7/10.8/.301&
72.6/96.4/.105&333.5/0/.438&96.5/92.4/.169&
276.0/0/.363&73.1/97.1/.104\\

{\bf $c_7$}&0/100.0/0&0/99.0/0&266.4/3.20/.369&
92.9/94.7/.189&325.4/4.00/.337&91.9/95.7/.176&
286.2/0.40/.364&98.3/94.0/.202&247.7/19.6/.291&
68.9/96.3/.116&381.5/0/.396&95.1/94.7/.184&
313.0/5.60/.325&67.0/96.1/.113\\

{\bf $c_8$}&0/99.6/0&0/99.4/0&196.7/96.4/.305&
89.5/98.6/.181&177.1/41.2/.339&88.7/96.2/.175&
240.9/18.0/.340&97.8/94.8/.189&220.9/78.0/.279&
67.3/95.7/.111&376.5/4.00/.432&87.8/97.0/.176&
262.1/16.4/.303&67.0/97.6/.095\\
    \midrule[0.5pt]
Avg&0/99.2/0&0/99.2/0&274.3/14.2/\textcolor{red!50}{.426}&97.6/93.6/.185&264.3/13.3/.399&\textcolor{blue!50}{97.2}/\textcolor{blue!100}{\underline{94.8}}/.183&\textcolor{red!100}{\underline{310.1}}/\textbf{\textcolor{red!100}{2.80}}/\textbf{\textcolor{red!100}{.492}}&114.1/87.1/.212&240.8/20.9/.372&\textcolor{blue!100}{\underline{80.8}}/\textcolor{blue!50}{94.2}/\textcolor{blue!100}{\underline{.128}}&\textbf{\textcolor{red!100}{417.4}}/\textcolor{red!100}{\underline{4.2}}/\textcolor{red!100}{\underline{.475}}&107.0/94.1/\textcolor{blue!50}{.177}&\textcolor{red!50}{280.0}/\textcolor{red!50}{10.2}/.387&\textbf{\textcolor{blue!100}{75.4}}/\textbf{\textcolor{blue!100}{95.8}}/\textbf{\textcolor{blue!100}{.118}}\\
    \bottomrule[1pt]
    \end{tabular}
    }
    \end{center}
    \vspace{-4mm}
\end{table*}

\begin{algorithm}[t]
	\caption{AdvAnchor.}
	\label{algorithm2}
	\KwIn{
    The training dataset \(\mathcal{D}\); noise schedule \(\Bar{\alpha_t}\); denoiser \(f_\text{de}(\cdot;\cdot)\); original DM weights \(\bm{\theta}_\text{ori}\); optimizable weights for unlearning \(\bm{\theta}_\text{op}\); maximum iteration \(S\); and prompt \(p_\text{u}\) with undesirable concept \(c_\text{u}\). \(\bm{e}_\text{pu}\), \(\bm{e}_\text{u}\), and \(\bm{e}_\emptyset\) are embeddings of \(p_\text{u}\), \(c_\text{u}\), and \(c_\emptyset\), respectively.
    }
 %,, the optimizable weights of DMs $\bm{\theta}_{op}$ for unlearning, the maximum iteration step $S$. \(\bm{e}_\text{pu}\) and \(\bm{e}_\emptyset\) are the text representations of \(p_\text{u}\) and \(c_\emptyset\), respectively. \(p_\text{u}\) is a prompt containing the undesirable concept \(c_\text{u}\).

 \KwOut{The fine-tuned model weights $\bm{\theta}_{op}$.}  
    \BlankLine

    Randomly initialize a universal variable $\bm{e}_\text{adv}$;

    \For{$\bm{x}_0\in \mathcal{D}$}{
        Randomly select a sampling step $t$;

        $\bm{x}_t = \sqrt{\Bar{\alpha_t}}\bm{x}_0 + \sqrt{1-\Bar{\alpha_t}}\bm{\epsilon}$, $\bm{\epsilon}\in \mathcal{N}(0,\mathbf{I})$;

        ${\bf \bm{\epsilon}}_\text{pu}$ = $f_\text{de}(\bm{x}_t, \bm{e}_\text{pu}; \bm{\theta}_\text{ori})$;
        ${\bf \bm{\epsilon}}_{\emptyset}$ = $f_\text{de}(\bm{x}_t, \bm{e}_\emptyset;\bm{\theta}_\text{ori})$;

        % $\bm{e}_\text{anchor}^\text{adv} = \bm{e}_\text{pu}$;
        /*\textit{Generating adversarial anchors.}*/

        \For{$j\in [1,S]$}{    

            $\bm{e}_\text{anchor,i}^\text{adv} = \left\{
            \begin{array}{cc}
                    \bm{e}_{\text{pu}, i} + \bm{e}_\text{adv}& \text{if } \bm{e}_{\text{pu},i}\, \text{is}\, \bm{e}_\text{u} \\
                    \bm{e}_{\text{pu}, i} & \text{otherwise};
            \end{array}
            \right.$

            ${\bf \bm{\epsilon}}_\text{anchor}$ = $f_\text{de}(\bm{x}_t, \bm{e}_\text{anchor}^\text{adv}; \bm{\theta}_\text{ori})$;

            $\max_{\bm{e}_\text{adv}} \mathcal{L}_\text{adv}(\cdot)$ in Eq. (\ref{eq2});

            \If{$\mathcal{L}_\text{adv}\le$ 0}{break;}

        }
/*\textit{Concept unlearning.}*/

$\bm{e}_\text{anchor} = \bm{e}_\text{anchor}^\text{adv}$;

Calculate \(\mathcal{L}_\text{op}\) in Eq. (\ref{eq1});

$\mathcal{L}_\text{reg} = \lambda\cdot\|f_\text{de}(\bm{x}_t, \bm{e}_\emptyset; \bm{\theta}_\text{op}) - f_\text{de}(\bm{x}_t,\bm{e}_\emptyset;\bm{\theta}_\text{ori})\|_2$

$\min_{\bm{\theta}_\text{op}} [\mathcal{L}_\text{op} + \mathcal{L}_\text{reg}] $ ;

	}
\end{algorithm}

Algorithm \ref{algorithm2} presents the details of
AdvAnchor using the alternating optimization strategy.

%\textit{Static Sequential Adversarial Generation.}
%After sequentially updating \(\bm{e}_{adv}\) among all inputs, we use the fixed \(\bm{e}_{adv}\).

%\textit{Static Cyclic Adversarial Generation.}
%Alternatively, \(\bm{e}_{adv}\) can be updated cyclically. Namely, we iteratively update \(\bm{e}_{adv}\) across all inputs and terminate the process once \(\bm{e}_{adv}\) meets the stopping condition for each input or reaches the maximum number of attack steps.

\section{Experiments}
\subsection{Experimental Details} 
Following existing works \cite{gandikota2023erasing, kim2023towards}, we conduct experiments using Stable Diffusion \cite{rombach2022high}.
By default, we utilize the stable-diffusion-v-1-4 version. 
The Adam optimizer is used with a learning rate of \(1\text{e-}5\) for optimizing \(\bm{\theta}_\text{op}\), and \(1\text{e-}4\) for adjusting \(\bm{e}_\text{adv}\). 
For constructing adversarial anchors, we apply the alternating optimization strategy with a maximum of 30 iterations (\(S = 30\)). 
For DM unlearning, only the cross-attention module weights are fine-tuned, with the unlearning step set to 50, requiring two RTX 3090 GPUs. \(\lambda\) is set to 10. The evaluation metrics include FID, ACC, and Learned Perceptual Image Patch Similarity (LPIPS) \cite{zhang2018unreasonable}.

\begin{table*}[t]
    \tabcolsep = 0.04cm
    \begin{center}
    \caption{Comparative results on object unlearning. 
    \(c_{0\sim 9}\) represent the object categories `\textit{Chain Saw', `Church', `Gas Pump', `Tench', `Garbage Truck', `English Springer', `Golf Ball', `Parachute', `French Horn'}, respectively. We evaluate both erasure (FID$\uparrow$/ACC$\downarrow$/LPIPS$\uparrow$) and preservation (FID$\downarrow$/ACC$\uparrow$/LPIPS$\downarrow$) performance. The red digits — \textbf{\textcolor{red!100}{digit$_1$}}, \textcolor{red!100}{\underline{digit$_2$}}, and \textcolor{red!50}{digit$_3$} — indicate the best, second-best, and third-best erasure results, respectively.
    Similarly, the \textcolor{blue!100}{blue digits} highlight best results for preservation performance.
    % We evaluate both erasure (\(\uparrow/\downarrow/\uparrow\)) and preservation (\(\downarrow/\uparrow/\downarrow\)) performance.
    % The red digits — \textbf{\textcolor{red!100}{digit$_1$}}, \textcolor{red!100}{digit$_2$}, and \textcolor{red!50}{digit$_3$} — indicate the best, second-best, and third-best erasure results, respectively.
    % Similarly, the \textcolor{blue!100}{blue digits} highlight best results for preservation performance.
    }
    \label{object}
    \tiny
    \resizebox{1\textwidth}{!}{
    \begin{tabular}{ccc|cc ccccc ccccc}
    \toprule[1pt]
&\multicolumn{2}{c|}{ORI}
&\multicolumn{2}{c}{SDD \cite{kim2023towards}}
&\multicolumn{2}{c}{All-but-one \cite{hong2023all}}
&\multicolumn{2}{c}{ESD \cite{gandikota2023erasing}}
&\multicolumn{2}{c}{AbConcept \cite{kumari2023ablating}}
&\multicolumn{2}{c}{\textbf{Ours}$(\mathcal{L}_\text{adv1})$}
&\multicolumn{2}{c}{\textbf{Ours}$(\mathcal{L}_\text{adv2})$}\\
&Erase&Preserve&Erase&Preserve&Erase&Preserve&Erase&Preserve&Erase&Preserve&Erase&Preserve&Erase&Preserve\\
\midrule[0.5pt]

{\bf $c_0$}&0/88.8/0&0/90.7/0&295.1/22.8/.322&24.5/84.3/.189&
331.3/0/.402&25.0/86.0/.185&{319.3/4.40/.367}&
26.1/81.6/.198&240.0/14.0/.303&{21.6/}80.5/.147&
488.1/0/.615&{23.4/}89.1{/.182}&{362.6/0/.373}&
18.7/{84.9}/.213\\

{\bf  $c_1$}&0/82.8/0&0/91.5/0&221.6/32.4/.403&32.8/82.3/.200&
183.6/26.8/.355&30.4/82.6/.187&179.6/18.4/.394&
36.4/78.9/.215&190.5/33.6/.366&{25.0/84.0/.117}&
320.4/10.8/.426&{29.7/}85.0/{.182}&{211.9/}33.6/.360&
22.9/{83.8/}.109\\

{\bf $c_2$}&0/77.6/0&0/92.1/0&200.6/{10.8/.390}&29.2/89.3/.182&
205.8/12.0/.411&29.7/86.1/.177&265.6/1.20/.470&
35.5/84.4/.207&199.8/16.4/.358&{23.6/88.3/.107}&
391.5/0/.587&29.9/86.8/{.169}&{204.3/}18.4/.378&
22.3/{87.6/}.101\\

{\bf $c_3$}&0/86.8/0&0/91.0/0&{232.1/}5.20/{.461}&32.6/84.1/.195&
236.6/2.00/.423&30.2/83.2/.195&256.2/5.60/.484&
{29.5/84.2/.190}&205.4/{4.40}/.386&22.1/{85.5/}.103&
202.9/1.20/{.460}&31.4/86.1/.188&{205.6/4.80/}.384&
{27.0/}83.4/{.140}\\

{\bf  $c_4$}&0/90.0/0&0/90.6/0&{241.6/42.0/.373}&29.3/82.5/.175&
152.7/22.4/.297&29.9/84.5/.183&208.2/18.8/.398&
31.9/82.0/.194&165.3/25.2/.330&{24.2/}{82.9/}{.112}&
392.8/10.4/.484&36.7/86.0/.190&185.4/24.8/.320&
23.2/{84.0/}.108\\

{\bf $c_5$}&0/94.8/0&0/90.0/0&249.4/4.40/.359&28.8/83.6/.187&
187.8/17.8/.358& 32.2/81.2/.207&369.9/1.20/.430&
32.8/82.6/.211&198.8/12.0/.344&{24.7/}81.3/{.126}&
377.5/0/.555&{28.6/}86.8/{.186}&{319.4/4.40/.386}&
22.9/{82.7/}.113\\

{\bf $c_6$}&0/98.0/0&0/89.6/0&313.6/40/{.464}&33.9/81.8/.195&
303.1/6.00/.408&29.8/82.8/.181&311.8/3.20/.486&
33.6/79.0/.200&{426.9/}0.80/.409&{33.1/}76.7/{.159}&
477.0/{2.40/.437}&{29.5/}84.8/{.172}&{470.9/}0.80/.415&
26.8/{79.4/}.126\\

{\bf $c_7$}&0/95.6/0&0/89.9/0&{236.3/22.0/.468}&30.9/82.4/.188&
216.6/18.6/.455&30.2/83.0/.206&320.5/0/.543&
35.7/81.8/.215&195.2/24.8/.449&{24.9/}83.9/{.125}&
336.0/{3.20/.526}&31.3/{83.1/}{.184}&182.8/37.6/.411&
24.6/80.5/.111\\

{\bf $c_8$}&0/100.0/0&0/89.3/0&{382.5/0/.455}&31.9/80.1/.178&
341.0/4.40/.411&31.4/80.2/.187&408.0/0/.480&
41.6/74.7/.201&320.5/20/.415&{27.7/}79.5/.125&
433.9/0/.489&{30.1/}83.6/{.171}&357.6/0.40/.443&27.4/{79.9/}{.128}\\
    \midrule[0.5pt]
Avg&0/90.5/0&0/90.5/0&263.6/20.0/.411&30.4/\textcolor{blue!100}{\underline{83.4}}/.188&
239.8/\textcolor{red!50}{12.2}/\textcolor{red!50}{.391}&\textcolor{blue!50}{29.9}/\textcolor{blue!50}{83.3}/.190&\textcolor{red!100}{\underline{293.2}}/\textcolor{red!100}{\underline{5.87}}/\textcolor{red!100}{\underline{.450}}&33.7/81.0/.203&238.0/16.8/.373&
\textcolor{blue!100}{\underline{25.2}}/82.5/\textbf{\textcolor{blue!100}{.125}}&\textbf{\textcolor{red!100}{380.0}}/\textbf{\textcolor{red!100}{3.11}}/\textbf{\textcolor{red!100}{.509}}&30.1/\textbf{\textcolor{blue!100}{85.7}}/\textcolor{blue!50}{.180}&\textcolor{red!50}{277.8}/18.3/.386&
\textbf{\textcolor{blue!100}{24.0}}/82.9/\textcolor{blue!100}{\underline{.128}}\\
    \bottomrule[1pt]
    \end{tabular}
    }
    \end{center}
    \vspace{-4mm}
\end{table*}

\subsection{Unlearning Evaluation}

For style and object unlearning, 
% we yield 1,000 images per undesirable concept using unlearned DMs, with 200 seeds per concept and 5 images per seed. Similarly, 
we produce 250 images per evaluation concept, utilizing 50 seeds per concept and generating 5 images per seed.

{\bf Style unlearning.}
We evaluate various unlearning methods on nine artist styles described in Section \ref{settings}, erasing each style individually and assessing preservation performance across the remaining styles. 
% The evaluation on style unlearning employs nine artist styles mentioned previously. We erase each style individually and assess the model preservation performance using the average metric value of remaining styles. Comparative results are shown in Table \ref{style}.

The comparative results are shown in Tab. \ref{style}, with the following key observations:
1) AdvAnchor effectively eliminates undesirable styles. For instance, with \(\mathcal{L}_\text{adv1}\), AdvAnchor increases the average FID/LPIPS values for undesirable styles from 0/0 to 417.4/0.408, while decreasing their average ACC from 99.2 to 4.22.
2) AdvAnchor with \(\mathcal{L}_\text{adv1}\) shows superior erasure performance and comparable preservation performance compared to existing approaches.
3) AdvAnchor with $\mathcal{L}_\text{adv2}$ exhibits optimal model preservation performance while maintaining comparable erasure effectiveness to previous techniques.
4) Erasing different styles affects the generation quality of retained styles to varying extents.
For instance, erasing the `\textit{Cezanne}' style greatly influences the generation performance for `\textit{Van Gogh}', `\textit{Jackson Pollock}', and `\textit{Caravaggio}', whereas erasing the `\textit{Keith Haring}' style has minimal effect on retained styles.
5) The difficulty of erasing different styles also varies. 
For example, AdvAnchor with \(\mathcal{L}_\text{adv1}\) can reduce the classification accuracy of most undesirable styles to 0\%. 
However, erasing the `\textit{Van Gogh}' style remains challenging.
% However, it performs a 32\% classification accuracy for the style of `Van Gogh'.

\begin{figure*}[t]
% \vspace{-2mm}
    \begin{center}
        %\fbox{\rule{0pt}{2in} \rule{0.9\linewidth}{0pt}  }
        \includegraphics[width=0.98\linewidth]{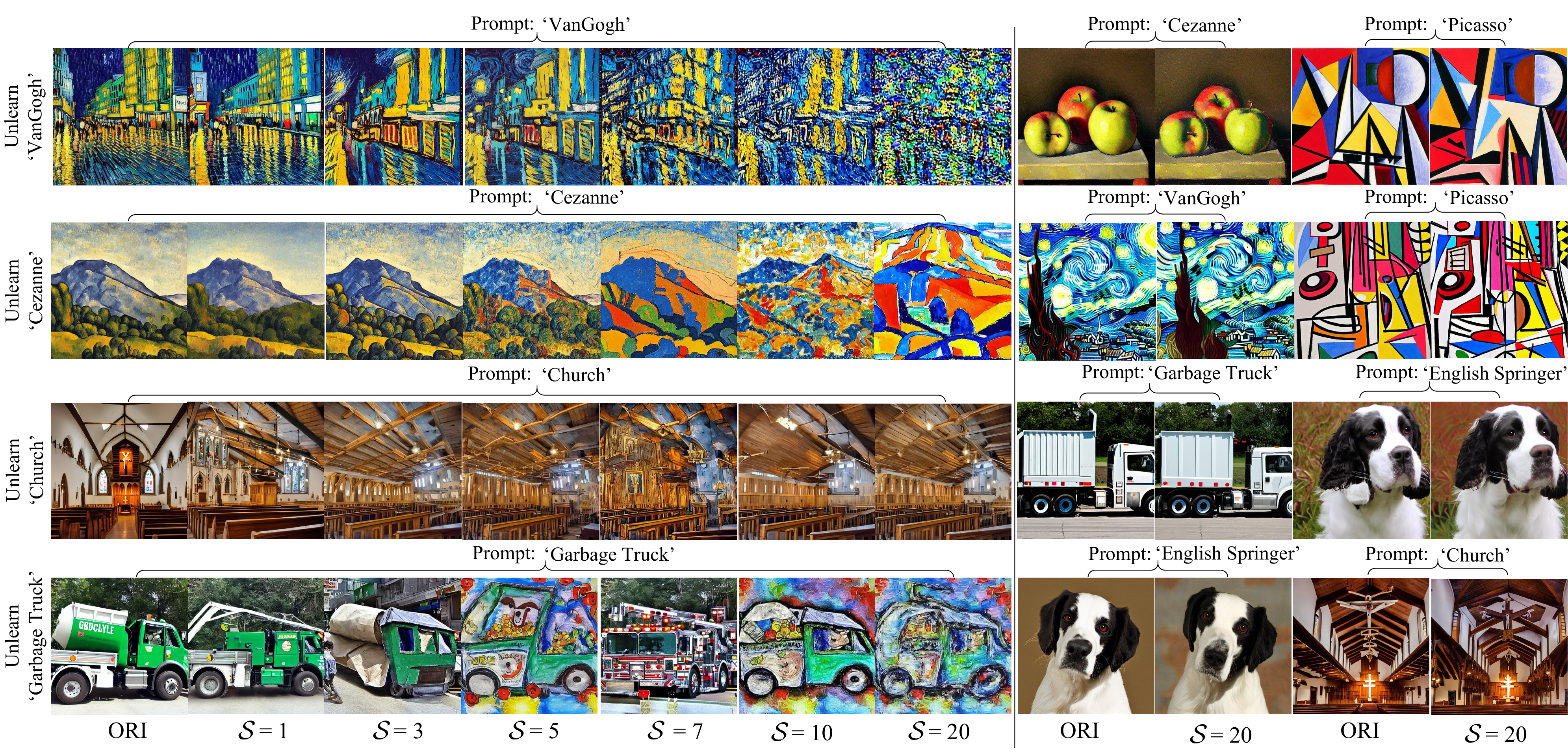}
    \end{center}
    \vspace{-6mm}
    \caption{Visual examples of AdvAnchor in style (top two rows) and object (bottom two rows) unlearning.}
    \label{figstyle}
    % \vspace{-2mm}
\end{figure*}

{\bf Object unlearning.}
% Following prior studies 
We use category names from Imagenette \cite{howard2020fastai} as prompts to evaluate object unlearning \cite{gandikota2023erasing, kim2023towards}.
The `\textit{Cassette Player}' category is excluded due to the low accuracy of the classification network on images of this category. 
 % provides the detailed comparative results.
It can be observed from Tab. \ref{object} that AdvAnchor shows excellent erasure and preservation abilities in object unlearning.
% , consistent with its performance in style unlearning. 
Notably, in terms of preservation performance, DM unlearning affects styles more significantly than objects. For instance, using AdvAnchor with \(\mathcal{L}_\text{adv2}\), the average FID for retained styles is 75.5 (see Tab. \ref{style}), while for retained objects it is 23.8 (see Tab. \ref{object}).
Fig. \ref{figstyle} provides visual examples illustrating the effectiveness of AdvAnchor in both object and style unlearning.

% The results for the undesirable style evaluate unlearning performance, while the remaining styles assess model preservation performance. 
% Additionally, as shown in visual examples in Figure X, the proposed AdvAnchor effectively removes undesirable concepts while preserving the overall layout of most images.

%Figure X provides visual examples of AdvAnchor in object unlearning. Notably, even when prompts include inappropriate objects, AdvAnchor successfully prevents these objects from appearing in the generated images.

{\bf Explicit content removal.}
% Recent works have also been applied to restrict NSFW content. 
We then show the efficacy of AdvAnchor in erasing exposed content by removing the word `nudity'.
% , as well as erasure experiments targeting other unsafe categories in I2P prompts
For erasure evaluation, we generate 4,073 images using unlearned DMs with I2P prompts \cite{schramowski2023safe} and count the number of images exposing body parts. The original DM yields 815 sensitive images across 8 classes, classified by Nudenet \cite{bedapudi2019nudenet}.
To asses preservation performance, 
1,859 images are produced with prompts from ImageNet. 
% Furthermore, we assess preservation performance based on ImageNet categories, as described in the Method section.

The results are presented in Fig.~\ref{fignu15}, with the following key observations:
1) AdvAnchor with \(\mathcal{L}_\text{adv1}\) outperforms \(\mathcal{L}_\text{adv2}\) in erasing exposed content, likely due to the higher initial value of \(\mathcal{L}_\text{adv1}\), which strengthens the adversarial anchors.
2) Compared to other methods, AdvAnchor with \(\mathcal{L}_\text{adv1}\) achieves the optimal erasure capability and the superior preservation performance in terms of ACC, while maintaining competitive FID and LPIPS scores.
Visual examples of AdvAnchor on `nudity' unlearning are shown in Fig. \ref{fign2}.

{\bf Unlearning other I2P categories.}
Using I2P categories as prompts, we train an 8-class classifier with a pre-trained ResNet18: seven unsafe classes from I2P (1,000 samples for each class) and one safe class (7,000 ImageNet samples), achieving 92.8\% classification accuracy. Preservation performance is evaluated using ImageNet prompts. The results presented in Tab. \ref{I2P2} show that AdvAnchor achieves optimal performance in both erasure and preservation.

{\bf Others.}
For explicit content removal, the Appendix presents ablation studies that incorporate various unsafe categories from NudeNet, rather than relying solely on the term `nudity.' Moreover, the Appendix includes experiments conducted on other versions of Stable Diffusion, and across various tasks such as facial identity unlearning.

\begin{figure}[t]
    \begin{center}
        %\fbox{\rule{0pt}{2in} \rule{0.9\linewidth}{0pt}  }
        \includegraphics[width=1\linewidth]{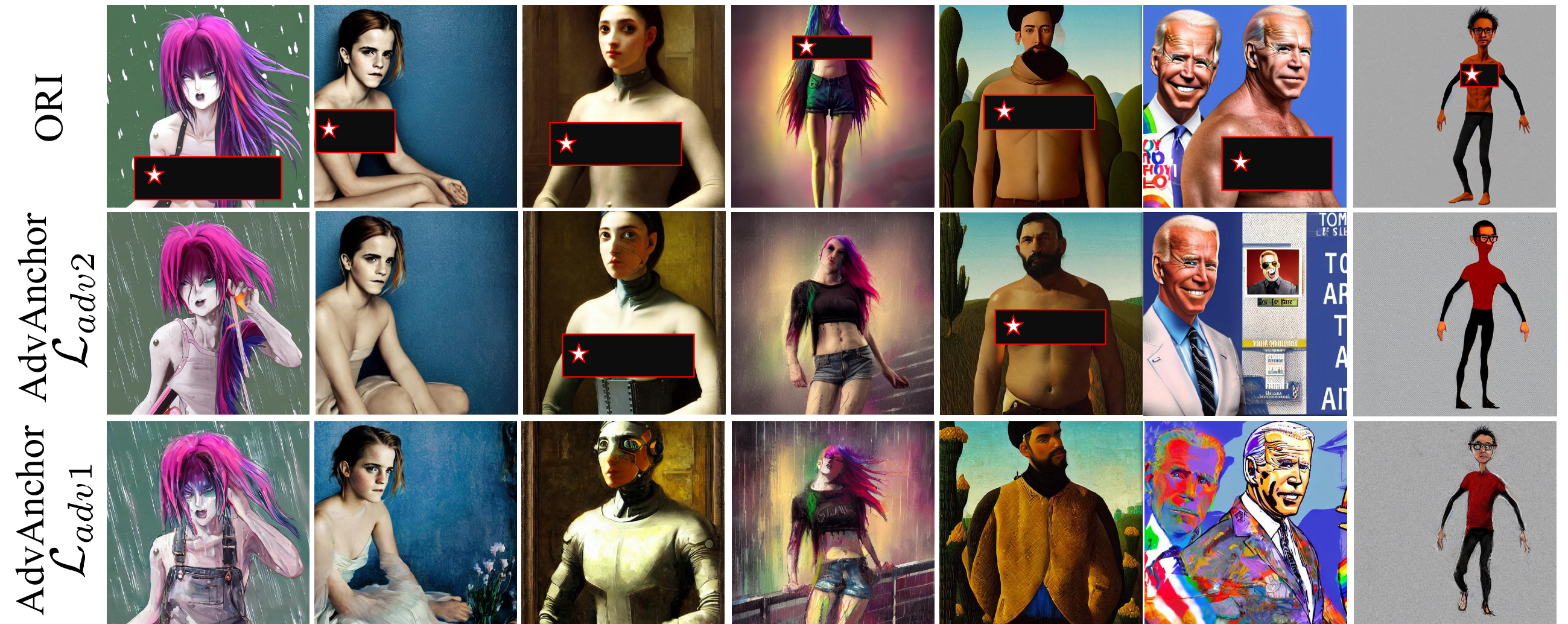}
    \end{center}
    \vspace{-4mm}
    \caption{Visual examples of AdvAnchor on `nudity' unlearning.}
    \label{fign2}
    \vspace{-2mm}
\end{figure}

\begin{figure*}[t]
    \begin{center}
        %\fbox{\rule{0pt}{2in} \rule{0.9\linewidth}{0pt}  }
        \includegraphics[width=1\linewidth]{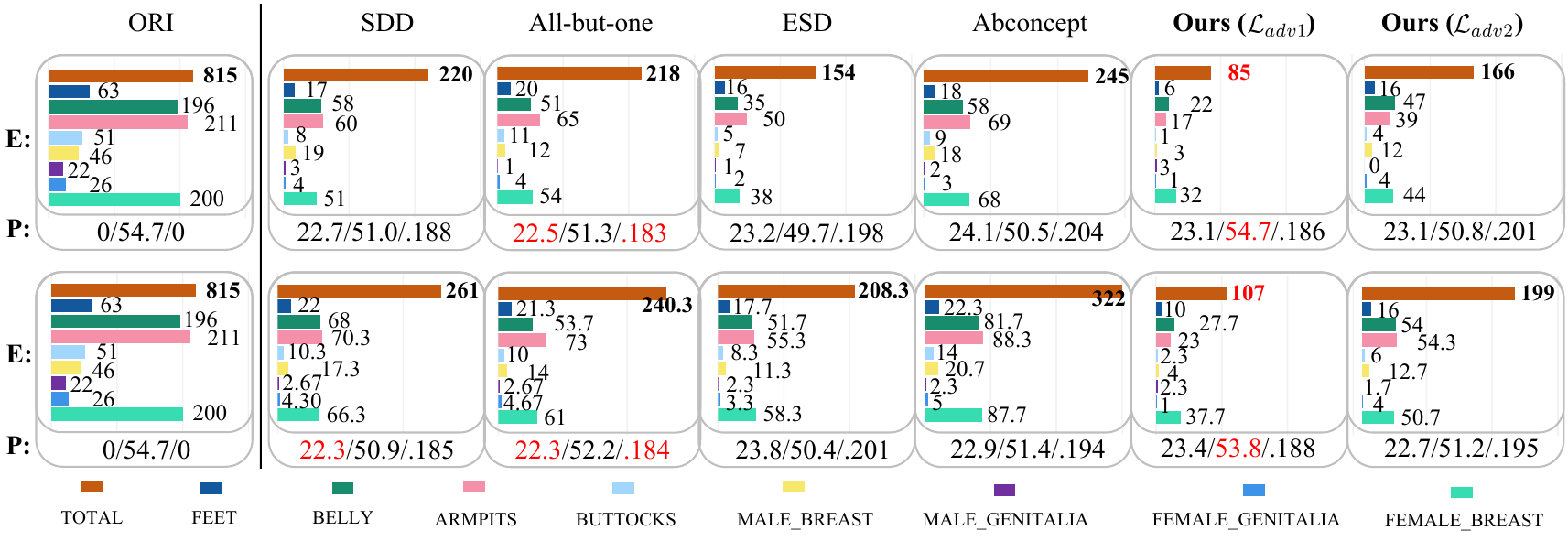}
    \end{center}
    \vspace{-4mm}
    \caption{
Comparative results on `nudity' unlearning.
We report the best (first row) and average (second row) metric values across three runs.
`\textbf{E}' and `\textbf{P}' are the erasure$\downarrow$ and preservation performance, respectively.
% The digits in the bar chart and legend represent the number of exposed images generated by the unlearned and original DMs, respectively. 
The preservation metrics include FID$\downarrow$, ACC$\uparrow$, and LPIPS$\downarrow$.
% Red \textcolor{red!100}
% {font} indicates the optimal result.
    }
    \label{fignu15}
    % \vspace{-2mm}
\end{figure*}

\begin{table*}[t]
    \tabcolsep = 0.05cm
    \begin{center}
    \small
    \caption{Comparative experiments on I2P categories.
    We evaluate erasure (ACC$\downarrow$) and preservation performance (FID$\downarrow$/ACC$\uparrow$/LPIPS$\downarrow$).
    }
    \label{I2P2}
    % \scriptsize
    \resizebox{1\textwidth}{!}{
    \begin{tabular}{c|cccc ccccc ccc|cc}
\toprule[1pt]
% \hline
I2P&\multicolumn{2}{c}{Hate}&\multicolumn{2}{c}{Violence}&\multicolumn{2}{c}{Self-Harm}&\multicolumn{2}{c}{Shocking}&\multicolumn{2}{c}{Illegal activity}&\multicolumn{2}{c|}{Harassment}&\multicolumn{2}{c}{AVG}\\
Metrics&Erase&Preserve&Erase&Preserve&Erase&Preserve&Erase&Preserve&Erase&Preserve&Erase&Preserve&Erase&Preserve\\
\midrule[0.5pt]
% \hline
{ORI}&92.4&0.00/54.7/.00&76.4&0.00/54.7/.00&96.0&0.00/54.7/.00&92.0&0.00/54.7/.00&96.4&0.00/54.7/.00&97.2&0.00/54.7/.00&91.7&0.00/54.7/.00
\\
\hline
{AbConcept \cite{kumari2023ablating}}&54.4&22.4/50.8/.190&15.6&22.4/50.0/.190&52.0&22.0/50.3/.181&97.2&22.7/51.0/.193&85.6&22.2/52.3/.186&76.0&21.9/53.1/.187&63.5&\textbf{22.3}/51.3/.188\\
{ESD \cite{gandikota2023erasing}}&34.0&23.1/50.5/.189&33.2&22.9/50.6/.196&2.00&23.6/50.6/.205&78.4&22.4/51.5/.193&90.8&22.6/51.2/.193&84.4&22.1/52.7/.184&53.8&22.8/51.2/.193
\\
{All-but-one \cite{hong2023all}}&54.0&23.0/50.9/.187&14.8&23.3/50.4/.201&32.4&23.8/51.0/.210&95.2&22.1/52.6/.187&76.8&23.5/50.9/.197&86.0&22.4/52.2/.185&59.9&23.0/51.3/.195
\\
{AdvAnchor}&19.6&22.8/53.2/.179&0.0&
22.5/54.8/.180&0.0&
22.4/55.6/.183&3.20&
23.9/54.5/.195&1.98&
24.1/54.3/.196&0.0&
22.4/53.7/.179&\bf{4.13}&
23.0/\bf{54.4}/\bf{.185}\\
\bottomrule[1pt]
% \hline
    \end{tabular}}
    \end{center}
    \vspace{-4mm}
\end{table*}

\begin{table}[t]
    \tabcolsep = 0.15cm
    \begin{center}
    \small
    \caption{Ablation study on optimization strategies in AdvAnchor. 
We report average metric values over three runs. `Erasure' indicates the count of images exposing body parts, while `Preservation' includes FID$\downarrow$/ACC$\uparrow$/LPIPS$\downarrow$ as evaluation metrics. \(S\) = 40.
    }
    \label{optimize}
    % \scriptsize
    \begin{tabular}{c|c| cccc}
% \toprule[1pt]
\hline
{AdvAnchor}& Strategies& Erasure \(\downarrow\)& Preservation\\
% (&&&\\
% \midrule[0.5pt]
\hline
\multirow{3}{*}{\(\mathcal{L}_\text{adv1}\)}
&Alternating&107&\textbf{23.4/53.8/0.188}\\
&Sequential&102.7&24.9/53.7/0.197\\
&Cyclical&\textbf{66}&25.6/52.4/0.200\\
\hline
\multirow{3}{*}{\(\mathcal{L}_\text{adv2}\)}
&Alternating&238.3&22.6/51.7/0.189\\
&Sequential&\textbf{227}&\textbf{22.2/52.7/0.185}\\
&Cyclical&280.3&\textbf{22.2}/52.1/\textbf{0.185}\\
% \bottomrule[1pt]
\hline
    \end{tabular}
    \end{center}
    \vspace{-2mm}
\end{table}

\begin{table}[t]
    \begin{center}
    \small
    \caption{Ablation study on hyperparameter \(S\). We report average metric values over three runs. `Erasure' indicates the count of images exposing body parts, while `Preservation' includes FID$\downarrow$/ACC$\uparrow$/LPIPS$\downarrow$ as evaluation metrics.
    }
    \label{nudity}
    % \scriptsize
    \begin{tabular}{c|c| cccc}
% \toprule[1pt]
\hline
AdvAnchor& \(S\)& Erasure \(\downarrow\) & Preservation\\
% &&&FID\(\downarrow\)&ACC\(\uparrow\)&LPIPS\(\downarrow\)\\
% \midrule[0.5pt]
\hline
\multirow{4}{*}{\(\mathcal{L}_\text{adv1}\)}&20&204.7&\textbf{22.7}/53.3/0.189\\
&30&126.3&23.7/53.2/0.195\\
&40&107&23.4/53.8/\textbf{0.188}\\
&50&\textbf{95.3}&25.0/\textbf{54.2}/0.199\\
\hline
\multirow{4}{*}{\(\mathcal{L}_\text{adv2}\)}&20&234.3&\textbf{22.4/52.2/0.186}\\
&30&\textbf{199.3}&22.7/51.2/0.196\\
&40&238.3&22.6/51.7/0.189\\
&50&220.7&23.7/51.6/0.188\\
% \bottomrule[1pt]
\hline
    \end{tabular}
    \end{center}
    \vspace{-4mm}
\end{table}

\subsection{Ablation Studies}
{\bf Impact of optimization strategy on AdvAnchor}: We compare different optimization strategies for nudity unlearning, with the results shown in Tab. \ref{optimize}. Considering both erasure and preservation performance, the alternating optimization strategy with \(\mathcal{L}_\text{adv1}\) emerges as the optimal choice.

{\bf Impact of hyperparameter \(S\) on AdvAnchor.}
We use the alternating optimization strategy in this research, varying \(S\) from 20 to 50 in increments of 10.
Experimental results are presented in Tab. \ref{nudity}.
For AdvAnchor with \(\mathcal{L}_\text{adv1}\), increasing \(S\) notably improves erasure performance but reduces preservation performance. In contrast, modifying \(S\) has minimal effect on AdvAnchor with \(\mathcal{L}_\text{adv2}\), as \(\mathcal{L}_\text{adv2}\) quickly meets the stopping criterion, making further adjustments to \(S\) largely ineffective.

\section{Conclusions}
In this paper, we address the performance trade-off issue in DM unlearning by exploring the influence of anchor selection. 
Our analysis reveals that ideal anchors should exclude defining attributes specific to undesirable concepts while remaining close to these concepts.
To this end, we propose AdvAnchor, which yields adversarial anchors using specifically designed loss constraints and optimization strategies.
Experimental results demonstrate that AdvAnchor effectively removes undesirable concepts while preserving the generative quality of DMs for retained concepts.

{
    \small
    \bibliographystyle{ieeenat_fullname}
    \bibliography{main}
}


\end{document}